\documentclass[11pt,a4paper]{article}
\usepackage[hyperref]{hypertext}
\usepackage{times}
\usepackage{latexsym}
\usepackage{amsfonts,amssymb,amsmath}
\usepackage{bm}
\usepackage[normalem]{ulem}

\usepackage{makecell}
\usepackage{microtype}

\usepackage{graphicx}
\graphicspath{{figures/}}

\usepackage{booktabs}
\usepackage{multirow}

\aclfinaltrue

\title{HyperText: Endowing FastText with Hyperbolic Geometry}

\author{Yudong Zhu\quad Di Zhou\quad Jinghui Xiao\quad Xin Jiang\quad Xiao Chen\quad Qun Liu\\
	Huawei Noah's Ark Lab\\
	\texttt{\{zhuyudong3,zhoudi7,xiaojinghui4,Jiang.Xin,} \\
	\texttt{chen.xiao2,qun.liu\}@huawei.com} \\}

\date{}

\begin{document}
\maketitle
\begin{abstract}
	Natural language data exhibit tree-like hierarchical structures such as the hypernym-hyponym relations in WordNet. FastText, as the state-of-the-art text classifier based on shallow neural network in Euclidean space, may not model such hierarchies precisely with limited representation capacity.  Considering that hyperbolic space is naturally suitable for modeling tree-like hierarchical data, we propose a new model named HyperText\footnote{\url{https://github.com/huawei-noah/Pretrained-Language-Model/tree/master/HyperText}}
for efficient text classification by endowing FastText with hyperbolic geometry. Empirically, we show that HyperText outperforms FastText on a range of text classification tasks with much reduced parameters.
\end{abstract}

\section{Introduction}
%Hierarchical structures are ubiquitous in a variety of research fields such as Biology, Social Networks, Knowledge Graphs. In particular,
FastText \cite{joulin2016bag} is a simple and efficient neural network for text classification, which achieves comparable performance to many deep models like char-CNN~\cite{zhang2015character} and VDCNN~\cite{conneau2016very}, with a much lower computational cost in training and inference. However, natural language data exhibit tree-like hierarchies in several respects \cite{dhingra2018hypertextembedding} such as the taxonomy of WordNet. In Euclidean space the representation capacity of a model is strictly bounded by the number of parameters. Thus, conventional shallow neural networks (e.g., FastText) may not represent tree-like hierarchies efficiently given limited model complexity.

Fortunately, hyperbolic space is naturally suitable for modeling the tree-like hierarchical data. Theoretically, hyperbolic space can be viewed as a continuous analogue of trees, and it can easily embed trees with arbitrarily low distortion \cite{Krioukov2010}. Experimentally,  \citet{nickel2017poincare} first used the Poincar\'{e} ball model to embed hierarchical data into hyperbolic space and achieved promising results on learning word embeddings in WordNet.
\begin{figure}[tt] 
	\setlength{\abovecaptionskip}{0.cm}
	\setlength{\belowcaptionskip}{-0.56cm}
	\centering
	\vspace{-0.15cm}
	\includegraphics[ width=0.5\textwidth,height=2.45in]{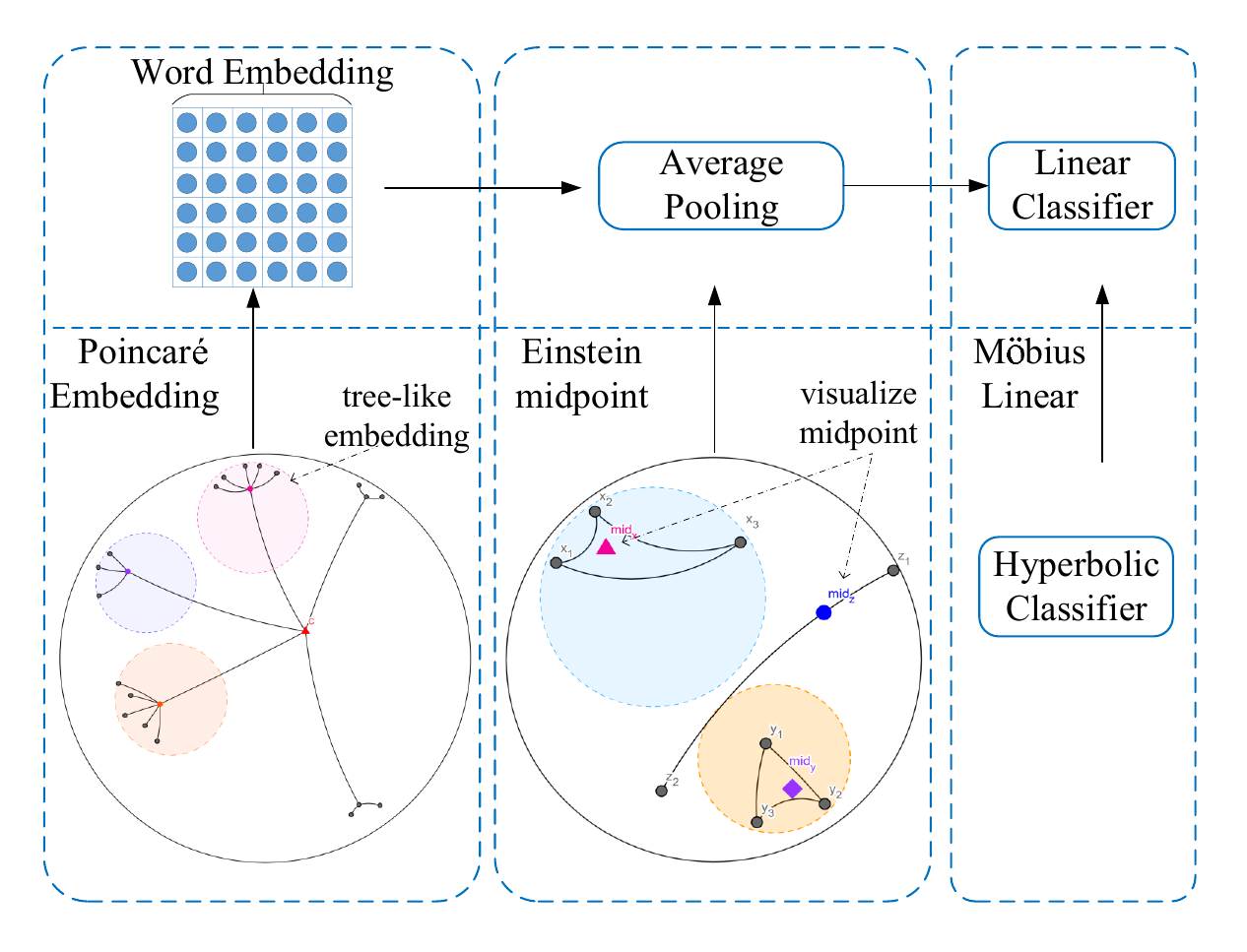}
	\caption{The architecture comparison of FastText (upper) and HyperText (lower).}
	\label{Model_Architecture}
\end{figure}

Inspired by their work, we propose HyperText for text classification by endowing FastText with hyperbolic geometry. We base our method on the Poincar\'{e} ball model of hyperbolic space. Specifically, we exploit the Poincar\'{e} ball embedding of words or ngrams to capture the latent hierarchies in natural language sentences. Further, we use the Einstein midpoint~\cite{HyperAttention} as the pooling method to emphasize semantically specific words which usually contain more information but occur less frequently than general ones~\cite{dhingra2018hypertextembedding}. Finally, we employ M\"obius linear transformation~\cite{ganea2018hyperbolic} to play the part of the hyperbolic classifier. We evaluate the performance of our approach on text classification task using ten standard datasets. We observe HyperText	\footnote{\url{https://gitee.com/mindspore/models/tree/master/research/nlp/hypertext}} outperforms FastText on eight of them. Besides, HyperText is much more parameter-efficient. Across different tasks, only $17\%\sim50\%$ parameters of FastText are needed for HyperText to achieve comparable performance. Meanwhile, the computational cost of our model increases moderately ($2.6$x in inference time) over FastText.

\section{Method}
\label{sec:length}

\subsection{Overview}
Figure \ref{Model_Architecture} illustrates the connection and distinction between FastText and HyperText. The differences of the model architecture are three-fold: First, the input token in HyperText is embedded using hyperbolic geometry, specifically the Poincar\'{e} ball model, instead of Euclidean geometry. Second, Einstein midpoint is adopted in the pooling layer so as to emphasize semantically specific words. Last, the M\"{o}bius linear transformation is chosen as the prediction layer. Besides, the Riemannian optimization is employed in training HyperText.

\subsection{Poincaré Embedding Layer}
There are several optional models of hyperbolic space such as the Poincar{\'e} ball model, the Hyperboloid model and the Klein model, which offer different affordances for computation. In HyperText, we choose the Poincar\'e ball model to embed the input words and ngrams so as to better exploit the latent hierarchical structure in text. The Poincar\'e ball model of hyperbolic space corresponds to the Riemannian manifold which is defined as follow:
\begin{equation}
\setlength{\abovedisplayskip}{3.5pt}
\setlength{\belowdisplayskip}{3.5pt}
\mathbb{P}^d= (\mathcal{B}^d,g_x),\label{poincareball}
\end{equation}
where $\mathcal{B}^d=\{\bm{x} \in\mathbb{R}^d \big|\left\|\bm{x}\right\|<1\} $ is an open $d$-dimensional unit ball ( $\left\|\bm{\cdot}\right\|$ denotes the Euclidean norm) and $g_x$ is the Riemannian metric tensor.
\begin{equation}
\setlength{\abovedisplayskip}{3.5pt}
\setlength{\belowdisplayskip}{3.5pt}
g_x= \lambda_x^2g^E,\label{hypermetric}
\end{equation}
where $\lambda_x=\frac{2}{1-\left\|\bm{x}\right\|^2}$ is the conformal factor, $g^E=\mathbf{I}_d$ denotes the Euclidean metric tensor. While performing back-propagation, we use the Riemannian gradients to update the Poincar\'e embedding. The Riemannian gradients are computed by rescaling the Euclidean gradients:
\begin{equation}
\setlength{\abovedisplayskip}{3.5pt}
\setlength{\belowdisplayskip}{3.5pt}
%Rgradf(\bm{x})= \frac{1}{1-\lambda_x^2}\bigtriangledown_Ef(\bm{x})\label{hypermetric}
\bigtriangledown_R f(\bm{x})= \frac{1}{1-\lambda_x^2}\bigtriangledown_Ef(\bm{x}).\label{hypermetric}
\end{equation}

Since ngrams retain the sequence order information, given a text sequence $S=\{w_i\}_{i=1}^m$, we embed all the words and ngrams into the Poincaré ball, denoted as $X=\{\bm{x}_i\}_{i=1}^{k}$, where $\bm{x}_i \in \mathcal{B}^d$.

\subsection{Einstein midpoint Pooling Layer}
Average pooling is a normal way to summarize features as in FastText. In Euclidean space, the average pooling is
\begin{equation}
\setlength{\abovedisplayskip}{3.5pt}
\setlength{\belowdisplayskip}{3.5pt}
%\mathrm{Avg}(\bm{x_1},\bm{x_2},...,\bm{x_n})
\bar{\bm{u}}= \frac{\sum_{i=1}^{k}\bm{x_i}}{k}.\label{avg}
\end{equation}
To extend the average pooling operation to the hyperbolic space, we adopt a weighted midpoint method called the Einstein midpoint \cite{HyperAttention}. In the $d$-dimensional Klein model $\mathbb{K}^d$, the Einstein midpoint takes the weighted average of embeddings, which is given by:
\begin{equation}
\setlength{\abovedisplayskip}{3.5pt}
\setlength{\belowdisplayskip}{3.5pt}
%\mathrm{HyperAvg}(\bm{x_1},\bm{x_2},...,\bm{x_n})
\bar{\bm{m}}_\mathbb{K}= \frac{\sum_{i=1}^{k} \gamma_i\bm{x_i}}{\sum_{i=1}^{k} \gamma_i}, \bm{x_i}\in\mathbb{K}^d\label{HyperAvg},
\end{equation}
where $\gamma_i=\frac{1}{\sqrt{1-\left\|\bm{x_i}\right\|^2}}$ are the Lorentz factors. %and $c$ is a hyper-parameter that denotes the curvature of hyperbolic spaces.
However, our embedding layer is based on the Poincar\'e model rather than the Klein model, which means we can't directly compute the Einstein midpoints using Equation (\ref{HyperAvg}).  Nevertheless, the various models commonly used for hyperbolic geometry  are isomorphic, which means  we can first project the input embedding to the Klein model, execute the Einstein midpoint pooling, and then project results back to the Poincar\'{e} model.

The transition formulas between the Poincar\'{e} and Klein models are as follow:
\begin{equation}
\setlength{\abovedisplayskip}{3.5pt}
\setlength{\belowdisplayskip}{0pt}
\bm{x}_\mathbb{K}=\frac{2\bm{x}_\mathbb{P}}{1+\left\|\bm{x}_\mathbb{P}\right\|^2},\label{p2k}
\end{equation}
\begin{equation}
\setlength{\belowdisplayskip}{1.5pt}
\bar{\bm{m}}_\mathbb{P}=\frac{\bar{\bm{m}}_\mathbb{K}}{1+\sqrt{1-\left\|\bar{\bm{m}}_\mathbb{K}\right\|^2}},\label{k2p}
\end{equation}
where $\bm{x}_{\mathbb{P}}$ and $\bm{x}_{\mathbb{K}}$ respectively denote token embeddings in the Poincar\'{e} and Klein models. $\bar{\bm{m}}_{\mathbb{P}}$ and $\bar{\bm{m}}_{\mathbb{K}}$ are the Einstein midpoint pooling vectors in the Poincar\'{e} and Klein models.
It should be noted that points near the boundary of the Poincar\'{e} ball get larger weights in the Einstein midpoint formula. These points (tokens) are regarded to be more representative, which can provide salient information for the text classification task \cite{dhingra2018hypertextembedding}.
%At this time, we can calculate the Lorentz factor $\gamma_i$ of every word vector $u_i$ in Klein space. $\gamma_i$ actually represents a weight value, indicating the importance of the word in the text sequence. Thus the output of Einstein midpoint pooling m is the weighted sum of each word vector $u_i$. The calculation method is as shown in Equation (3)-(4).
%In addition, since the m is located in the Klein space now, in order to facilitate the calculation of the next layer, it needs to be mapped back to the Poincaré space. Therefore, the final output of the Einstein midpoint pooling layer is p.
\subsection{M\"obius Linear Layer}
The M\"obius linear transformation is an analogue of linear mapping in Euclidean neural networks. We use the M\"obius linear to combine features outputted by the pooling layer and complete the classification task, which takes the form:
\begin{equation}
\setlength{\abovedisplayskip}{3.8pt}
\setlength{\belowdisplayskip}{3.8pt}
%f^{\otimes_c}(\jx{\bar{\bm{m}}})
\bm{o}=\bm{M} {\otimes}\bar{\bm{m}}_{\mathbb{P}}\ \oplus\ \bm{b},\label{mobiuslinear}
\end{equation}where $\otimes$ and $\oplus$ denote the M\"obius matrix multiplication and M\"obius addition  defined as follows \cite{ganea2018hyperbolic}:

\begin{small}
	\setlength{\abovedisplayskip}{0.0pt}
	\begin{equation}
	\setlength{\abovedisplayskip}{0.0pt}
	\setlength{\belowdisplayskip}{1.5pt}
	\bm{M}{\otimes}\bm{x}=(1/\sqrt{c})\tanh\big(\frac{\left\|\bm{M}\bm{x}\right\|}{\left\|\bm{x}\right\|}\tanh^{-1}(\sqrt{c}\left\|\bm{x}\right\|)\big)\frac{\bm{M}\bm{x}}{\left\|\bm{M}\bm{x}\right\|},\notag
	\end{equation}
	
	\begin{equation}
	\setlength{\abovedisplayskip}{3.5pt}
	\setlength{\belowdisplayskip}{1.1pt}
	\bm{x}\oplus\bm{b}=\frac{(1+ 2c \langle \bm{x}, b \rangle + c{\left\|\bm{b}\right\|}^2  )\bm{x} + (1-c{\left\|\bm{x}\right\|}^2)\bm{b}}   {1+ 2c \langle \bm{x},\bm{b} \rangle +c^2 {\left\|\bm{x}\right\|}^2 {\left\|\bm{b}\right\|}^2}.\notag
	\end{equation}
\end{small}
\begin{table*}[t]	
	
	\small
	\resizebox{\textwidth}{12mm}{
		\begin{tabular}{lcccccccccc}
			\hline
			\textbf{Model}      & \textbf{AG}   & \textbf{Sogou} & \textbf{DBP}  & \textbf{Yelp P.} & \textbf{Yelp F.} & \textbf{Yah. A.} & \textbf{Amz. F.} & \textbf{Amz. P.} & \textbf{TNEWS} & \textbf{IFYTEK} \\ \hline
			FastText            & 92.5          & 96.8          & 98.6          & 95.7          & 63.9          & 72.3          & 60.2          & 94.6          & 54.6          & 54.0          \\ 
			VDCNN               & 91.3          & 96.8          & 98.7          & 95.7          & \textbf{64.7} & 73.4          & \textbf{63.0} & \textbf{95.7} & 54.8          & \textbf{55.4} \\ 
			DistilBERT(1-layer)$^*$ & 92.9          & -             & \textbf{99.0} & 91.6          & 58.6          & 74.9          & 59.5          & -             & -             & -             \\ 
			FastBERT(speed=0.8) & 92.5          & -             & \textbf{99.0} & 94.3          & 60.7          & \textbf{75.0} & 61.7          & -             & -             & -             \\ \hline
			HyperText           & \textbf{93.2} & \textbf{97.3} & 98.5          & \textbf{96.1} & 64.6          & 74.3          & 60.1          & 94.6          & \textbf{55.9} & 55.2          \\ \hline
	\end{tabular}}
\caption{\label{main dataset experiment results}
	Accuracy(\%) of different models. $^*$The results of DistilBERT are cited from \citet{fastbert2020}
}
\vspace{-0.2cm}
\end{table*}where $\bm{M} \in\mathbb{R}^{d \times n} $ denotes the weight matrix, and $n$ denotes the number of class; $\bm{b} \in \mathbb{R}^{ n} $ is the bias vector and $c$ is a hyper-parameter that denotes the curvature of hyperbolic spaces.
%$$f=Mobius_multi(p,W) \eqno{(4)}$$
In order to obtain the categorical probability $\hat{\bm{y}}$ , a softmax layer is used after the M\"obius linear layer.
$$
\setlength{\abovedisplayskip}{3.5pt}
\setlength{\belowdisplayskip}{3.5pt}
\hat{\bm{y}}=\mathrm{softmax}(\bm{o})\eqno{(9)}
$$
\subsection{Model Optimization}
This paper uses the cross-entropy loss function  for the multi-class classification task:
%$$L=cross_entropy(y,label) \eqno{(7)}$$
$$
\setlength{\abovedisplayskip}{3.5pt}
\setlength{\belowdisplayskip}{3.5pt}
L=-\frac{1}{N}\sum_{i=1}^{N} \bm{y} \cdot \log(\hat{\bm{y}}), \eqno{(10)}
$$
where $N$ is the number of training examples, and $\bm{y}$ is the one-hot representation of ground-truth labels. For training, we use the Riemannian optimizer \cite{becigneul2018riemannian} which is more accurate for the hyperbolic models. We refer the reader to the original paper for more details.

\section{Experiments}
\subsection{Experimental setup}
%\cite{joulin2016bag}   \cite{CLUEbenchmark}
\paragraph{Datasets} To make a comprehensive comparison with FastText, we choose the same eight datasets as in \citet{joulin2016bag} in our experiments. Also, we add two Chinese text classification datasets from Chinese CLUE \cite{CLUEbenchmark}, which are presumably more challenging. We summarize the statistics of datasets used in our experiments in Table \ref{Dataset statistics}.
	\paragraph{Hyperparameters} Follow \citet{joulin2016bag}, we set the embedding dimension as $10$ for first eight datasets in Table \ref{main dataset experiment results}. On TNEWS and IFLYTEK datasets, we use $200$-dimension and $300$-dimension embeddings respectively. The learning rate is selected on a validation set from $\{0.001,0.05, 0.01, 0.015\}$. In addition, we use PKUSEG tool~\cite{pkuseg}  for Chinese word segmentation.
	\begin{table}[t]
	\centering
	\small
	\begin{tabular}{l|ccc}
		\hline
		\textbf{Dataset} & \textbf{\#Classes} & \textbf{\#Train} & \textbf{\#Test} \\ \hline
		\textit{AG}               & 4                  & 120,000          & 7,600           \\
		\textit{Sogou}            & 5                  & 450,000          & 60,000          \\ 
		\textit{DBP}              & 14                 & 560,000          & 70,000          \\ 
		\textit{Yelp P.}          & 2                  & 560,000          & 38,000          \\ 
		\textit{Yelp F.}         & 5                  & 650,000          & 50,000          \\
		\textit{Yah. A.}          & 10                 & 1,400,000        & 60,000          \\ 
		\textit{Amz. F.}          & 5                  & 3,000,000        & 650,000         \\ 
		\textit{Amz. P.}          & 2                  & 3,600,000        & 400,000         \\ 
		\textit{TNEWS}            & 15                 & 53,360           & 10,000          \\ 
		\textit{IFLYTEK}          & 119                & 12,133           & 2,599           \\ \hline
	\end{tabular}
	\caption{\label{Dataset statistics}
		Dataset statistics
		\vspace{-0.6cm}
	}
\end{table}
	\subsection{Experimental Results}
	
	\paragraph{Comparison with FastText and deep models}
	The results of our experiments are displayed in Table \ref{main dataset experiment results}. Our proposed HyperText model outperforms FastText on eight out of ten datasets, and the accuracy of HyperText is $0.7\%$ higher than FastText on average. In addition, from the results, we observe that HyperText works significantly better than FastText on the datasets with more label categories, such as Yah.A., TNEWS and IFLYTEK. This arguably confirms our hypothesis that HyperText can better model the hierarchical relationships of the underlying data and extract more discriminative features for classification. Moreover, HyperText outperforms DistilBERT\cite{distilbert2019} and FastBERT\cite{fastbert2020} which are two distilled versions of BERT. And HyperText achieves comparable performance to the very deep convolutional network (VDCNN)~\cite{conneau2016very} which consists of 29 convolutional layers. From the results, we can see that HyperText has better or comparable classification accuracy than these deep models while requiring several orders of magnitude less computation.

%which  The average error rate was reduced from 15.68 to 15.16, relatively reduced by 3.25\%. This shows that our HyperText model based on Poincaré space does have stronger feature extraction and combination capabilities, and could extract some more complex features and better model the hierarchical relationship between words and n-grams. In addition, our HyperText just use 2-gram now. if we use more n-grams and bigger hidden size, the error rate will be further decreased by 0.05 $ \sim $ 0.5 on these 8 datasets.
\begin{figure*}[t] 
	\centering 
	\setlength{\abovecaptionskip}{0.1cm}
	\includegraphics[ width=0.97\textwidth,height=2.2in]{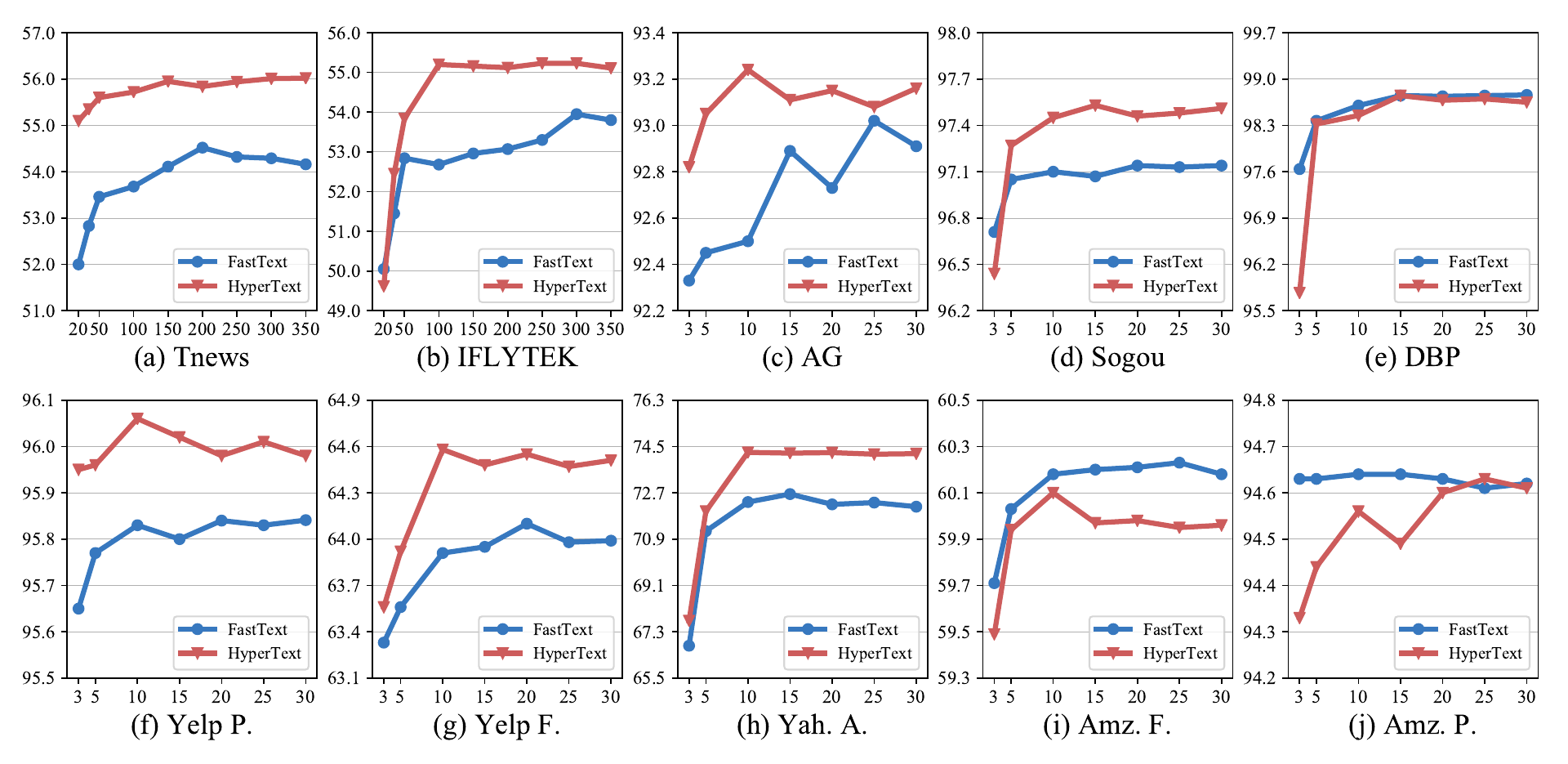}
	\caption{Accuracy vs Embedding dimension. The $x$-axis represents the embedding dimension, while the $y$-axis represents the accuracy.}
	\label{Accuracy vs Embedding Dimension} 
	\vspace{-0.5cm}
\end{figure*}
\paragraph{Embedding Dimension}
Since the input embeddings account for more than 90\% model parameters, we investigate the impact of dimension of input embedding on the classification accuracy. The experimental results are presented in Figure \ref{Accuracy vs Embedding Dimension}. As we can see, on most tasks HyperText performs consistently better than FastText in various dimension settings. In particular, on IFLYTEK and TNEWS datasets, HyperText with $50$-dimension respectively achieves better performance to FastText with $300$-dimension and $200$-dimension. On other eight less challenging datasets, the experiments are conducted in the low-dimensional settings and HyperText often requires less dimensions to achieve the optimal performance in general. It verifies that thanks to the ability to capture the internal structure of the text, the hyperbolic model is more parameter efficient than its Euclidean competitor.

%Since the amount of embedding parameters accounts for more than 90\% of the total parameters of the model, reducing the dimension of embedding will greatly reduce the model size. So we analyzed the impact of embedding dimension for the performance of different models on the IFLYTEK dataset. The chart of accuracy score changes with embedding dimension is shown in Figure 2. It can be seen from the figure that HyperText is basically above FastText, and there is a large merge. This shows that HyperText is significantly better than FastText again. In addation, the figure shows that HyperText, justly using 50-dimensions embedding, achieves performance comparable to FastText with 300-dimension embedding. This also shows that HyperText does significantly reduce the model size, and could achieve comparable performance with FastText that justly using 1/6 of the parameters of the FastText.

\begin{figure}[h] 
	\centering 
	\setlength{\abovecaptionskip}{0.1cm}
	\includegraphics[ width=0.45\textwidth,height=2.1in]{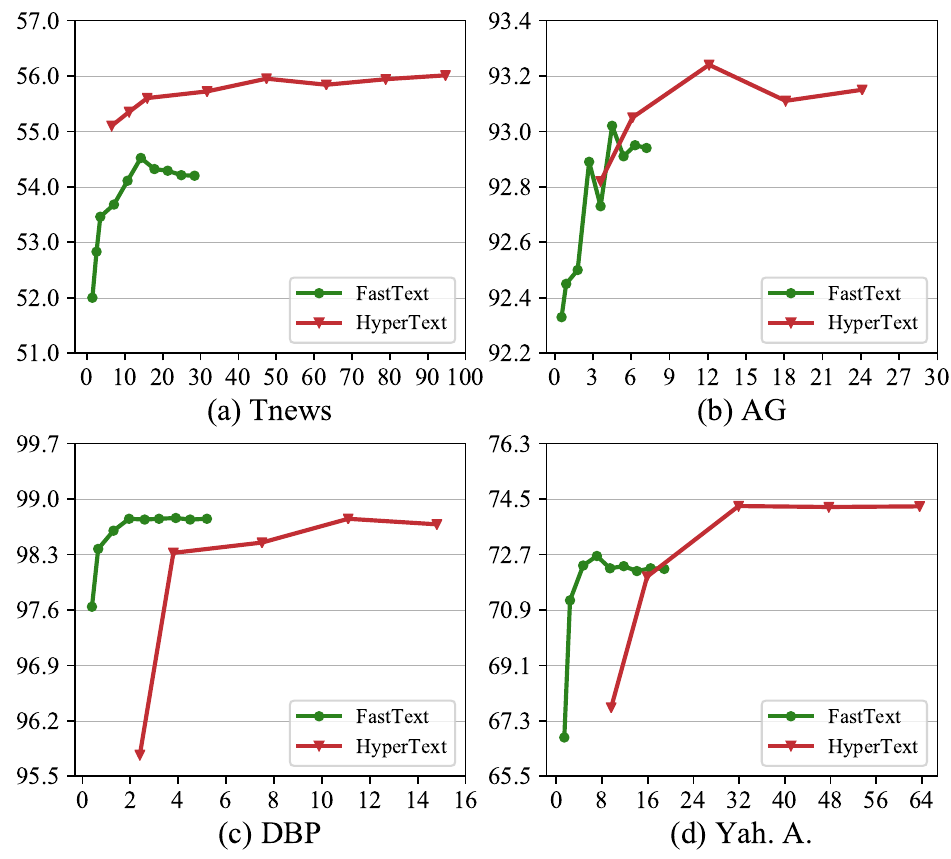} 
	\caption{FLOPs($\times 10^3$) vs Accuracy(\%) under different dimensions. The $x$-axis represents the FLOPs, while the $y$-axis represents the accuracy. Different points represent different embedding dimensions} 
	\label{FLOPs vs Accuracy} 
	\vspace{-0.4cm}
\end{figure}

\paragraph{Computation in Inference} FastText is well-known for its fast inference speed. We compare the FLOPs versus accuracy under different dimensions in Figure \ref{FLOPs vs Accuracy}. Due to the additional non-linear operations, HyperText generally requires more ($4.5\sim6.7$x) computations compared to FastText with the same dimension. But since HyperText is more parameter efficient, when constrained on the same level of FLOPs, HyperText mostly performs better than FastText on the classification accuracy. Besides, the FLOPs level of VDCNN is $10^5$ higher than HyperText and FastText.

\paragraph{Ablation study}
We conduct the ablation study to figure out the contribution of different layers. The results on several datasets are present in Table \ref{ablation study}. Note that whenever we replace a hyperbolic layer with its counterpart in Euclidean geometry, the model performs worse. The results show that all the hyperbolic layers (Poincar\'{e} Embedding Layer, Einstein midpoint Pooling Layer and M\"{o}bius Linear Layer) are necessary to achieve the best performance.

%\sout{By putting all layer into hyperbolic geometry, we get the best performance. In addition,  because these Poincaré Embedding Layer and Einstein midpoint pooling Layer are interdependent, and only using one of them, the model cannot be trained well, we justly analysis the result that replacing them together.}

%In order to analyze the contribution of different layers in HyperText, we only hyperbolic part layers every time. FastText is based on European space, which means that all layers are non-hyperbolic. HyperText\_Linear means that just last linear layer is based on hyperbolic space. The ablation results are shown in Table 4. As you can see from the table, HyperText\_Linear is 0.6 higher than FastText. This also shows that the Mobius linear layer does have stronger feature extraction and combination capabilities. In addition, HyperText has further improved compared to HyperText\_Linear, and the overall improvement is 1.1. That also verifies that the embedding layer based on Poincaré space and the Einstein midpoint pool layer could  better represent the hierarchical relationship between words and n-grams. Meanwhile, it could make different words have a larger margin in Poincaré space.
\begin{table}[h]
	\centering
	\small
	\begin{tabular}{lllll }
		\hline
		\textbf{Model} & \textbf{Yelp P.}& \textbf{AG}& \textbf{Yah.A.} & \textbf{TNEWS}  \\
		\hline
		
		HyperText & 96.1 & 93.2 &  74.3 &  55.9     \\
		\quad-PE\&EM & 95.9&  92.8 &  73.9 & 55.6 \\
		\quad-ML & 95.6&  92.3&  73.2  & 54.6 \\
		
		\hline
	\end{tabular}
	\caption{\label{ablation study}
		Ablation study of each components in HyperText (PE for Poincar\'{e} Embedding, EM for Einstein Midpoint, and ML for M\"{o}bius Linear layer).
	}
\vspace{-0.6cm}
\end{table}

\section{Related Work}
%FastText \cite{joulin2016bag}  has received extensive attention since it was proposed, especially in industry. FastText  is an efficient neural network based on Euclidean geometry, which achieves comparable performance to many deep models.  %Major drawbacks of FastText are high memory requirement and its inability to model the latent hierarchies of natural language.
Hyperbolic space can be regarded as a continuous version of tree, which makes it a natural choice to represent the hierarchical data \cite{nickel2017poincare,nickel2018poincare,desa2018}. Hyperbolic geometry has been applied to learning knowledge graph representations. HyperKG \cite{Hyperkg2019} extends TransE to the hyperbolic space, which obtains great improvement over TransE on WordNet dataset.  \citet{MURPgraphembedding2019} proposes MURP model which minimizes the hyperbolic distances between head and tail entities in the multi-relational graphs. Instead of using the hyperbolic distance, \citet{chami2019graphrotationembedding,chami2020graphembedding} uses the hyperbolic rotations and reflections to better model the rich kinds of relations in knowledge graphs. Specifically, the authors use the hyperbolic rotations to capture anti-symmetric relations and hyperbolic reflections to capture symmetric relations, and combine these operations together by the attention mechanism. It achieves significant improvement at low dimension.
Hyperbolic geometry is also applied in natural language data so as to exploit the latent hierarchies in the word sequences \cite{Tifrea2019poincareglove}.

Recently, many hyperbolic geometry based deep neural networks \cite{HyperAttention,ganea2018hyperbolic} achieve promising results, especially when the mount of parameters is limited. There are some applications based on hyperbolic geometry, such as question answering system \cite{hyperqa}, recommendation system \cite{hyperrecommendsystem2019} and  image embedding \cite{hyperimageembedding}.

\section{Conclusion}

We have shown that hyperbolic geometry can endow the shallow neural networks with the ability to capture the latent hierarchies in natural language. The empirical results indicate that HyperText consistently outperforms FastText on a variety of text classification tasks. On the other hand, HyperText requires much less parameters to retain performance on par with FastText, which means neural networks in hyperbolic space could have a stronger representation capacity.

%\section*{Acknowledgments}

%The acknowledgments should go immediately before the references. Do not number the acknowledgments section.
%Do not include this section when submitting your paper for review.

%\bibliography{anthology,hypertext}
%\bibliographystyle{acl_natbib}

\end{document}